\documentclass[10pt,twocolumn,letterpaper]{article}
\usepackage{cvpr}              

\usepackage{graphicx}
\usepackage{amsmath}
\usepackage{amssymb}
\usepackage{booktabs}
\usepackage{wrapfig}
\usepackage{caption}
\usepackage{subcaption}
\usepackage{algorithm}
\usepackage{algorithmic}
\usepackage{tabularx}
\usepackage{multirow}
\usepackage{booktabs}  
\usepackage{pifont}
\usepackage{tikz}
\usepackage[accsupp]{axessibility}

\newcommand{\cmark}{\ding{51}}%
\newcommand{\xmark}{\ding{55}}%

\usepackage[pagebackref,breaklinks,colorlinks]{hyperref}

\usepackage[capitalize]{cleveref}
\crefname{section}{Sec.}{Secs.}
\Crefname{section}{Section}{Sections}
\Crefname{table}{Table}{Tables}
\crefname{table}{Tab.}{Tabs.}

\def\cvprPaperID{9345} 
\def\confName{CVPR}
\def\confYear{2023}

\begin{document}

\title{Leveraging per Image-Token Consistency for Vision-Language Pre-training}

\author{Yunhao Gou$^{1,2}$\thanks{\, Work was done when the author interned at ByteDance AI Lab.}
\and
Tom Ko$^{3}$
\and
Hansi Yang$^{2}$
\and
James Kwok$^{2}$
\and
Yu Zhang$^{1, 4}$\thanks{\, The corresponding author.}
\and
Mingxuan Wang$^{3}$
\and
\\
$^{1}$Southern University of Science and Technology, $^{2}$Hong Kong University of Science and Technology \\
$^{3}$ByteDance AI Lab, $^{4}$Peng Cheng Laboratory \\
{\tt\small \{ygou, hyangbw\}@connect.ust.hk} \quad {\tt\small jamesk@cse.ust.hk} \\
{\tt\small \{tom.ko, wangmingxuan.89\}@bytedance.com} \quad
{\tt\small yu.zhang.ust@gmail.com} \\
}
\maketitle

\begin{abstract}
Most existing vision-language pre-training (VLP) approaches adopt cross-modal masked language modeling (CMLM) to learn vision-language associations.
However, we find that CMLM is insufficient for this purpose according to our observations: 
(1) Modality bias: a considerable amount of masked tokens in CMLM can be recovered with only the language information, ignoring the visual inputs.
(2) Under-utilization of the unmasked tokens: CMLM primarily focuses on the masked token but it cannot simultaneously leverage other tokens to learn vision-language associations.
To handle those limitations, we propose \textbf{EPIC} (l\textbf{E}veraging \textbf{P}er \textbf{I}mage-Token \textbf{C}onsistency for vision-language pre-training).
In \textbf{EPIC}, for each image-sentence pair, we mask tokens that are salient to the image (i.e., Saliency-based Masking Strategy) and replace them with alternatives sampled from a language model (i.e., Inconsistent Token Generation Procedure), and then the model is required to determine \emph{\textbf{for each token}} in the sentence whether it is consistent with the image (i.e., Image-Token Consistency Task).
The proposed EPIC method is easily combined with pre-training methods. Extensive experiments show that the combination of the EPIC method and state-of-the-art pre-training approaches, including ViLT, ALBEF, METER, and X-VLM, leads to significant improvements on downstream tasks. Our code is released at \url{https://github.com/gyhdog99/epic}
\end{abstract}

\begin{figure*}[t]
    \centering
    \includegraphics[width=\textwidth]{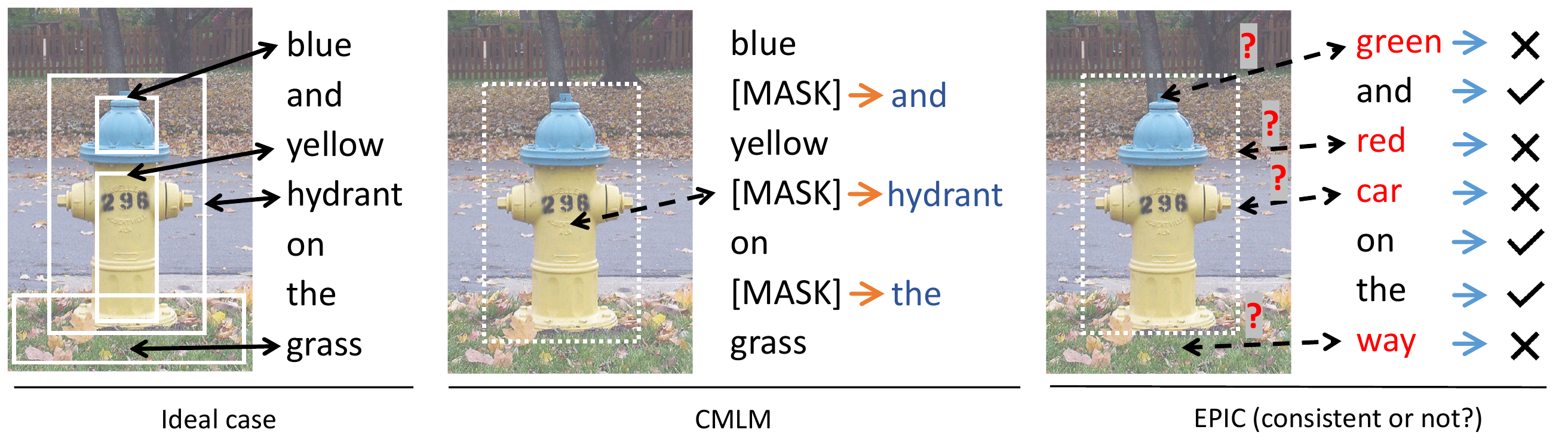}
\caption{Illustrations of vision-language association learning. \textbf{Ideal
case}: Fine-grained annotations (image regions and corresponding text tokens) are
given, we can learn explicit associations (solid lines); \textbf{CMLM}: Without
fine-grained annotations, we create supervision by masking, but this can be insufficient due to limited masking ratios and modality bias. \textbf{EPIC}: We find salient tokens and corrupt them to learn more associations. Both \textbf{CMLM} and \textbf{EPIC} learn implicit associations due to lack of region annotations.}
    \label{fig:intro}
\end{figure*}

\section{Introduction}
\label{sec1}
Vision-language pre-training (VLP) \cite{lu2019vilbert, chen2020uniter, su2019vl, wang2021simvlm, kim2021vilt, singh2022flava, yang2022vision} aims to learn multi-modal representations from large-scale image-text pairs. 
A pre-trained vision-language model (VLM) fine-tuned with only a small amount of labeled data has shown state-of-the-art performance in many downstream tasks such as visual question answering and image-text retrieval.

A primary concern in developing pre-training objectives for VLP models is how to learn better vision-language associations. 
In addition to coarse-grained approaches such as image-text matching/contrasting \cite{pmlr-v139-radford21a, Li2021AlignBF, jia2021scaling} that align concepts from two modalities at the sample level, fine-grained approaches such as cross-modal masked language/image modeling (CMLM/CMIM) \cite{tan-bansal-2019-lxmert, liu2021kd, li2020oscar} learn vision-language associations at the token-object level.
For example, Fig.~\ref{fig:intro} shows a picture paired with the sentence
``Blue and yellow hydrant on the grass''. When the word 
``hydrant''
is masked,
in order to correctly recover the token, the model has to find 
the actual object in the image 
and associate  it
with the word ``hydrant''.

While effective, CMLM is insufficient for learning vision-language associations
because of (1) modality bias; and (2) under-utilization of unmasked tokens.
In vision-language understanding,
modality bias 
refers to leveraging only one modality for training/inference and so cross-modal knowledge is not well explored \cite{niu2021counterfactual}.
We argue that modality bias exists in CMLM, and prevents the model from learning sufficient vision-language associations.
Specifically, in the CMLM task, we expect
to mask \emph{salient}\footnote{The definition of ``saliency'' is given in Sec. \ref{saliency}.} 
tokens
(such as ``blue'', ``yellow'', ``fire-hydrant'', and ``grass'') as shown in the left of Fig.~\ref{fig:intro}.
These tokens are informative for learning vision-language association because masking them enforces the model to find the answer from the visual modality.
However, 
in practice,
whether a token is salient is unknown 
as we only have access to image-sentence level annotations.
Given a fixed and relatively small masking ratio (typically 15\% in CMLM), we might end up masking tokens that are less informative.
For example, as shown in Fig.~\ref{fig:intro} (center), 
when ``the'' and ``and'' are masked,
the model can predict these masked tokens 
with only language information. This thus 
is a form of modality bias
as it circumvents using vision-language reasoning.
Therefore, the modality bias can make CMLM insufficient to learn vision-language associations.

Another source of insufficiency in CMLM comes from the under-utilization of unmasked tokens.
Similar to Masked Language Modeling (MLM) \cite{devlin2018bert} in language pre-training, the CMLM loss is computed over masked tokens rather than all tokens in the sentence.
As a result, learning of cross-modal association is possible only for the masked
tokens but not for the remaining unmasked ones.
For example, in Fig. \ref{fig:intro}, ideally, there are four associations
(shown in black arrows) between text tokens and the corresponding regions, while
there is only one association for CMLM.
Therefore, CMLM cannot leverage all tokens (including the unmasked ones) for learning vision-language associations.

To expedite the learning of cross-modal associations in VLP, we propose \textbf{EPIC} (l\textbf{E}veraging \textbf{P}er \textbf{I}mage-Token \textbf{C}onsistency for vision-language pre-training).
For each image-sentence pair, we mask tokens that are salient to the image
(Saliency-based Masking Strategy) and make them ``inconsistent''\footnote{A formal
definition of (in)consistency tokens will be provided in Sec. \ref{itc}.} with the
image by replacing them with alternatives from a BERT-like language model
(Inconsistent Token Generation Procedure). The model is then required to determine
whether 
\emph{\textbf{each token}} in the sentence 
is consistent with the image (Image-Token Consistency (ITC) Task).
As this masks salient tokens and applies a language model to generate inconsistent tokens from them, the model has to refer to the visual modality to determine whether a token is inconsistent.
Therefore, the modality bias problem can be alleviated.
Moreover, we can make better use of the unmasked tokens for learning vision-language association as the ITC task requires the model to determine whether \emph{\textbf{each token}} is consistent with the image.

The proposed EPIC method is easy to implement and widely applicable to a lot of vision-language model architectures. 
We demonstrate the effectiveness of our approach on various pre-training
approaches, including ViLT \cite{lu2019vilbert}, ALBEF \cite{Li2021AlignBF}, METER
\cite{dou2022meter}, and X-VLM \cite{xvlm}, and observe significant improvements on downstream tasks.
For example, on MSCOCO image-text retrieval, the proposed EPIC method achieves an absolute gain of 2.5\% and 4.7\% over METER and ViLT, respectively, in terms of the Recall@1 score.
On visual reasoning tasks (e.g., NLVR2), the proposed method improves over ALBEF by 1.8\% and \mbox{X-VLM} (the state-of-the-art within its model scale) by 1.3\%. 
The proposed method also allows better generalization of pre-training models.
For example, in zero-shot image-text retrieval, we improve X-VLM by 3.9\% (COCO) and ViLT by 9.9\% (Flickr30k).

\section{Related Work}

\noindent\textbf{Learning Vision-Language Associations.}
In vision-language pre-training, 
a contrastive objective is often used to
learn the 
coarse-grained 
associations between sentences and images 
\cite{pmlr-v139-radford21a, Li2021AlignBF, jia2021scaling, you2022learning, li2021supervision}.
However, models pre-trained with this objective give unsatisfactory results on
vision-language reasoning tasks (such as VQA \cite{antol2015vqa} and NLVR2
\cite{suhr2018corpus}) which require understanding fine-grained vision-language associations.
Cross-Modal Masked Language/Image Modeling (CMLM/CMIM) \cite{liu2021kd,
li2020oscar, bitton-etal-2021-data-efficient, lu2022cots} and its variants are
applied to learn fine-grained vision-language associations in a self-supervised
manner.
Besides CMLM/CMIM, one can 
further leverage annotated region-phrase (e.g., bounding boxes) image-text data to enable the model with more sophisticated vision-language reasoning abilities
\cite{xvlm, dou2022coarse, kamath2021mdetr, li2022grounded}.

\noindent\textbf{Cross-Modal Masked Language/Image Modeling.}
CMLM and CMIM are widely adopted as pre-training objectives in vision-language pre-training. 
They corrupt a token/patch of a sentence/image and train the model to reconstruct the original data with information from both modalities.
LXMERT \cite{tan-bansal-2019-lxmert} found that loading BERT into the text encoder
harms pre-training because the pre-trained BERT can have high CMLM accuracy and
thereby circumventing the cross-modal reasoning process.
Recently, similar to our observations, Bitton et al.
\cite{bitton-etal-2021-data-efficient} showed that roughly 50\% of the masked
tokens in CMLM are punctuation or stop-words, leading to sample inefficiency of CMLM.
They propose a rule-based masking strategy to mask object words, content words, or
words with high concreteness \cite{Brysbaert2014ConcretenessRF} in CMLM according
to different downstream tasks, expecting such words to be more related to the
visual input.
Such a strategy lacks generalization ability and cannot scale to more downstream tasks.

\noindent\textbf{Modality Bias in Vision-Language Understanding.}
In CMLM, predicting the masked token without referring to the visual modality can be seen as a consequence of modality bias towards language in vision-language understanding. 
Due to the existence of modality bias, the learning of vision-language associations is weakened.
Similarly, modality bias also happens in visual question answering (VQA) \cite{antol2015vqa} where the model is required to answer a question given a visual input. 
For example, simply answering ``tennis'' to the sport-related questions can achieve approximately 40\% accuracy \cite{niu2021counterfactual} on the VQA v1.0 dataset.
To reduce such a bias, CF-VQA \cite{niu2021counterfactual} proposes a counterfactual framework which directly subtracts the language-based predictions from the answers. 
In the proposed EPIC method, we mitigate the negative influence of language bias by allowing the model to learn more vision-language associations over a wider range of tokens.

\section{Empirical Analysis}
In this section, we empirically analyse the existence of modality bias and under-utilization of unmasked tokens in CMLM. 
Experimental
details 
are
in Appendix \ref{apx:emp}.
\label{sec2}

\begin{figure}
     \begin{subfigure}[b]{0.23\textwidth}
         \centering
         \includegraphics[width=\textwidth]{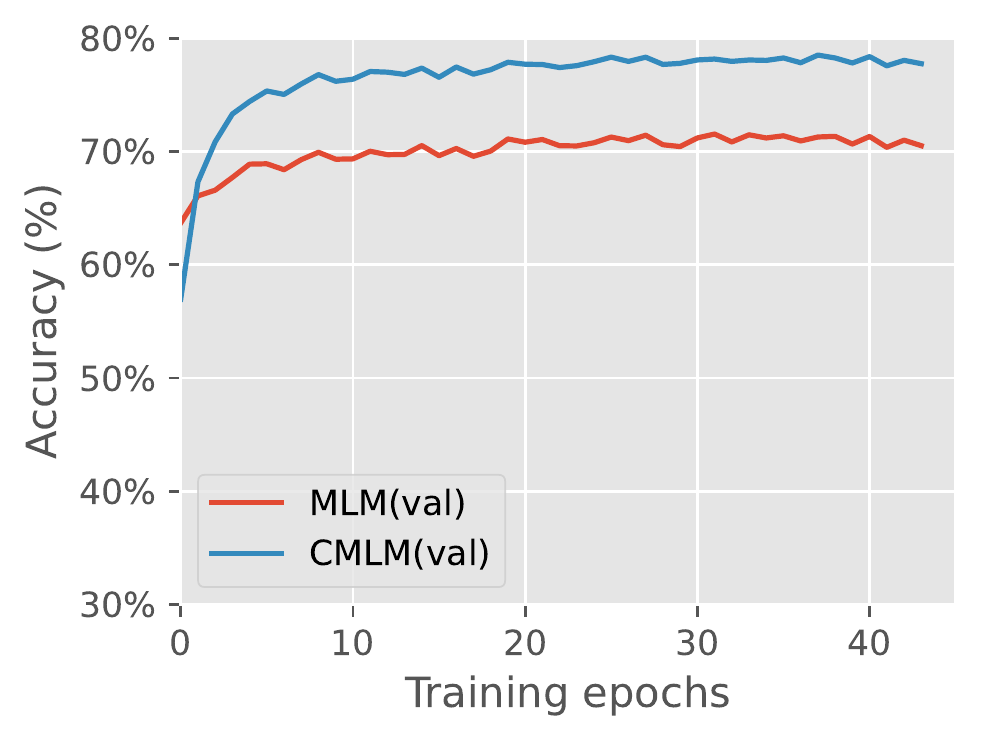}
         \caption{Accuracy gaps.}
         \label{fig:emp_bias_val}
     \end{subfigure}
     \hfill
     \begin{subfigure}[b]{0.23\textwidth}
         \centering
         \includegraphics[width=\textwidth]{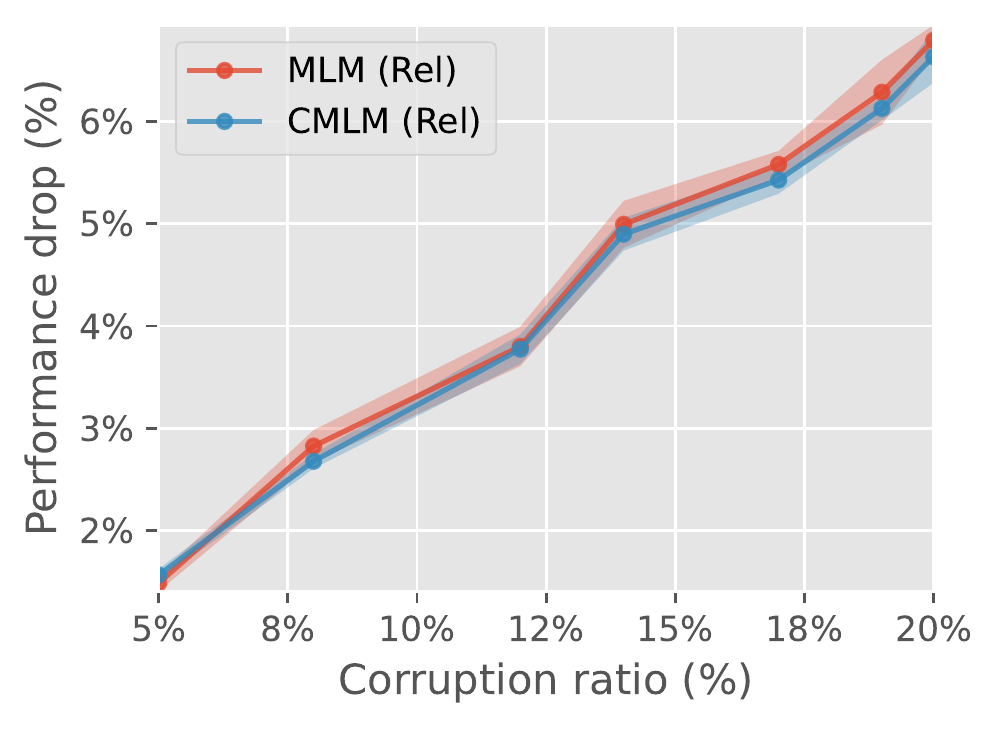}
         \caption{Accuracy drops.}
         \label{fig:emp_insen_rel}
     \end{subfigure}
     \hfill
     \caption{Left: CMLM/MLM accuracy of VLM/LM (y-axis) as the training process proceeds (x-axis). Right: relative accuracy drop of CMLM/MLM (y-xais) as we corrupt more unmasked tokens (x-axis). The colored bands indicate the variance of the results as each configuration is repeated for 5 times.}
\end{figure}

\noindent\textbf{Identifying Modality Bias.}
If there is a strong modality bias towards language in Cross-Modal Language
Modeling (CMLM), a pure language model (LM) that is blind to the visual inputs can achieve comparable accuracy on the Masked Language Modeling (MLM) task. 
Otherwise, the LM will be outperformed by the vision-language model (VLM) by a large margin.
Based on this, we run CMLM on a VLM and MLM on a LM, respectively.
They share the same input sentences and masked tokens, while the former additionally receives visual inputs. 
As shown in Fig. \ref{fig:emp_bias_val}, though using visual modality is helpful
to recover the original token (80\% Acc.), a pure LM can already achieve an accuracy of 70\%.
Therefore, a modality bias towards language does exist in the training of CMLM.

\noindent\textbf{Under-utilization of Unmasked Tokens.}
If the CMLM is weak at reasoning associations between images and \emph{unmasked}
tokens, replacing the unmasked ones with alternatives based on the language
context should not prevent the model from recovering the masked tokens; otherwise,
the performance of CMLM will drop significantly because the language context is inconsistent with the image. 
Therefore, we use a pre-trained VLM and LM to perform inference on CMLM and MLM,
respectively, under a corrupted context in which we randomly replace unmasked tokens by sampling from the MLM head of a BERT.
As illustrated in Fig. \ref{fig:emp_insen_rel}, both the VLM and LM suffer
performance deterioration under different corruption ratios.
However, even though the text contexts are inconsistent with the image after
corruption, the performance drop is less severe for the VLM as compared to the LM.
In fact, the performance curves for CMLM and MLM overlap in terms of relative performance drop.
Hence, we conclude that the CMLM is weak at learning associations between unmasked tokens and images.

\begin{figure*}[t]
    \centering
    \includegraphics[width=\textwidth]{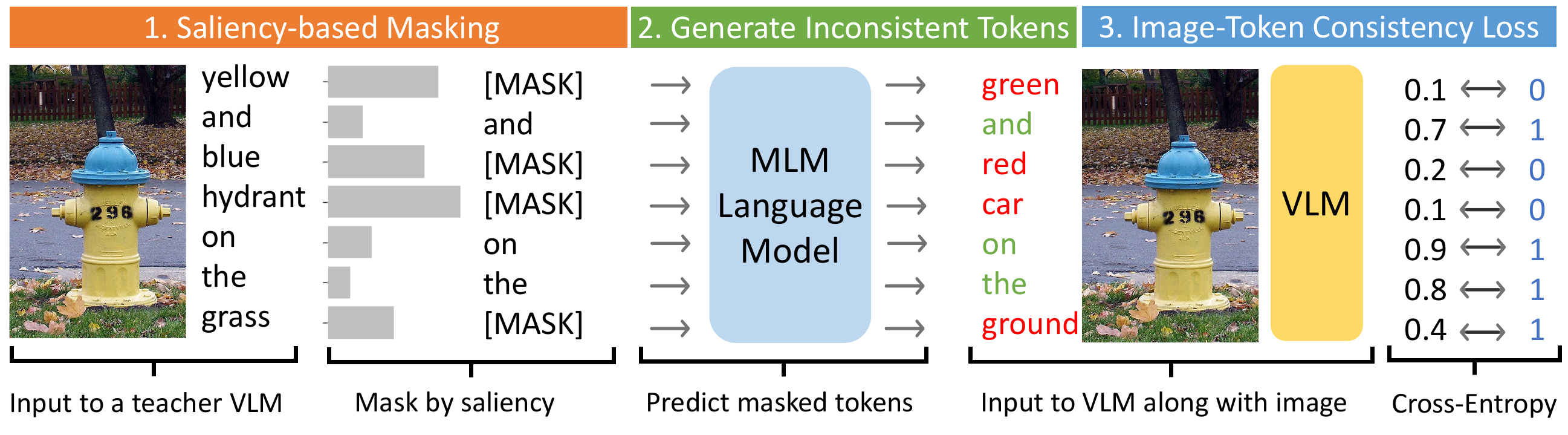}
    \caption{An overview of EPIC. The input image and sentence are input to a teacher VLM to obtain the saliency for each text token w.r.t the image. Then we mask accordingly and generate tokens inconsistent with the image though a language model (fine-tuning during pre-training). Finally we train the VLM to determine for each token whether it is inconsistent with the image (ITC).}
    \label{fig:overview}
\end{figure*}

\section{Methodology}

Sec.~\ref{sec:pre}
first introduces preliminaries of the proposed method.
We then 
introduce the three components of \textbf{EPIC}.
We formulate the ITC task and introduce the concept of token consistency w.r.t. an image in Sec.~\ref{itc}.
We design the inconsistent token generation procedure in Sec.~\ref{negative}, and the saliency-based masking strategy in Sec.~\ref{saliency}.
Fig. \ref{fig:overview} gives an overview of the proposed method.
The complete EPIC algorithm 
is shown 
in Appendix \ref{apx:alg}.

\subsection{Preliminaries}
\label{sec:pre}
\noindent\textbf{Pre-training Framework.}
In vision-language pre-training, we have access to parallel image-text data $\mathcal{D} = \left\{(\boldsymbol{w}_{i}, \boldsymbol{v}_{i})\right\}_{i=1}^{N} \sim P_{\mathcal{W, V}}$. 
Specifically, for any sentence $\boldsymbol{w}=\left[w_{1}, \ldots, w_{n}\right]$, there is a corresponding image $\boldsymbol{v}=\left[v_{1}, \ldots, v_{m}\right]$ in the form of grid-based or region-based features \cite{dou2022meter}.
Without loss of generality, we assume a vision-language model $f_{\text{VL}}(\cdot)$ in METER \cite{dou2022meter} of the following form:
Given a pair of sentence $\boldsymbol{w}$ and image $\boldsymbol{v}$, it first
extracts text features and visual features via a text encoder and vision encoder,
respectively.
The text and visual features are then fed into a multi-modal fusion module,
consisting of several layers of self-attention followed by cross-attention, to produce cross-modal representations for text and image.
Note that 
the proposed method is also applicable
to other architectures such as ViLT \cite{kim2021vilt}, ALBEF \cite{Li2021AlignBF}, and X-VLM \cite{xvlm}.

\subsection{Image-Token Consistency}
\label{itc}

First, we introduce the concept of consistency of a text token with respect to an image. 

\noindent\textbf{Consistency of Text Tokens w.r.t Images.}
In vision-language pre-training, we assume that the dataset is clean (without noise) and that we have access to the marginal distribution of the text corpus $P_{\mathcal{W}}$. For a pair of text $\boldsymbol{w}$ and image $\boldsymbol{v}$, a text token $w_{i}$ in $\boldsymbol{w}$ is ``consistent'' with the image given the remaining tokens, i.e., $P_{\mathcal{W, V}}(\boldsymbol{W} = \boldsymbol{w}, \boldsymbol{V} = \boldsymbol{v})$ is large.
If we replace a token $w_j$ with $w^{\prime}_{j}$ such that the new sentence
$\boldsymbol{\bar{w}}$ (1) does not match the semantic content of the image, i.e.,
$P_{\mathcal{W, V}}(\boldsymbol{W} = \boldsymbol{\bar{w}}, \boldsymbol{V} =
\boldsymbol{v})$ is small and (2) is linguistically proper, i.e.,
$P_{\mathcal{W}}(\boldsymbol{W} = \boldsymbol{\bar{w}})$ is large, then
$w^{\prime}_{j}$ is inconsistent with the image given the remaining tokens.

An example is shown in Fig.~\ref{fig:overview}. In the sentence ``Yellow and blue
hydrant on the grass'', the word ``hydrant'' is consistent with the image given
the remaining words. If we replace ``yellow'' with ``green'', it produces a
sentence that does not match the image, and so ``green`` is inconsistent.

\noindent\textbf{Task Formulation.}
Given an image $\boldsymbol{v}$ and a sentence $\boldsymbol{\bar{w}}$ with
inconsistent tokens at positions $\mathcal{T}$, the model has to determine whether each token is consistent with the image.
We will introduce the generation process of $\boldsymbol{\bar{w}}$ and $\mathcal{T}$ in Sec.~\ref{negative}.

The $\bar{\boldsymbol{w}}$ and $\boldsymbol{v}$ are fed into VLM to obtain the last-layer hidden representations 
$\left\{\bar{\boldsymbol{h}}^{\text{VL}}_{i}\right\}_{i=1}^{n}$
of $\bar{\boldsymbol{w}}$.
Let $\boldsymbol{\beta}$ be the weight vector for decision. The probability that
each token (from position $i=1$ to $n$) is consistent with the image is:
\begin{equation}
    \label{eq:sigmoid}
    D_{\text{ITC}}^{i} = \operatorname{sigmoid}\left(\boldsymbol{\beta}^{\intercal}\bar{\boldsymbol{h}}^{\text{VL}}_{i}\right).
\end{equation}
Therefore, the EPIC model minimizes the following binary classification loss:
\begin{equation}
    \label{eq:litc}
    \begin{split}
        \mathcal{L}_{\mathrm{ITC}} =
        -\sum_{i \not\in \mathcal{T}} \log D_{\mathrm{ITC}}^{i} 
        -\sum_{i \in \mathcal{T}} \log \left(1-D_{\mathrm{ITC}}^{i}\right).
    \end{split}
\end{equation}

\noindent\textbf{Leveraging Unmasked Tokens.}
It is easy to see that in Eq. \eqref{eq:litc}, the EPIC model is tasked to determine for all the tokens whether they are consistent with the image or not.
Therefore, the ITC task leverages more tokens for learning vision-language associations.

\noindent\textbf{Mitigating Modality Bias.}
In the second condition for inconsistent tokens, we require them to be linguistically proper for the sentence.
We argue that this is essential for the ITC task to alleviate the modality bias
problem because if the sentence after replacement $\boldsymbol{\bar{w}}$ is not
linguistically proper, then the VLM can identify the replaced tokens simply by using the language context only.
As a result, the VLM does not learn vision-language associations by decreasing $\mathcal{L}_{\mathrm{ITC}}$. Instead, it simply conducts language modeling.

\subsection{Generating Inconsistent Tokens}
\label{negative}

In this section, we discuss how to generate sentences $\boldsymbol{\bar{w}}$ with tokens that fulfill the two conditions of inconsistent tokens in Sec.~\ref{itc}. 
Strictly speaking, we have to model the distributions $P_{\mathcal{W, V}}$ and
$P_{\mathcal{W}}$ for generation, but this is hard. Instead, we propose to approximately generate inconsistent tokens with a BERT-like language model.

Assume that we have a set 
$\mathcal{M}$ 
of masking positions 
(we will discuss how to obtain $\mathcal{M}$ in Sec. \ref{saliency}). Given the original text sequence $\boldsymbol{w}$, we first mask the corresponding tokens by replacing them with the \texttt{[MASK]} symbol:
\begin{equation}
    \label{eq:replace2}
    \boldsymbol{w}^{\text{mask}} = \operatorname{MASK}\left(\boldsymbol{w},
	 \mathcal{M}\right).
\end{equation}
The resulting sentence $\boldsymbol{w}^{\text{mask}}$ is fed to an auxiliary BERT-like language model $f_{L}(\cdot)$ to obtain a sequence of contextual representations $\boldsymbol{H}^{\text{L}} = \left\{\boldsymbol{h}^{\text{L}}_{i}\right\}_{i=1}^{n}$.
We then obtain the probability of predicting a particular token $w_{k}$ as:
\begin{equation}
    \label{eq:mlm_prob}
    p_{\mathrm{MLM}}\left(w_{k} \mid \boldsymbol{h}^{\text{L}}_{i}\right)= \frac{\exp \left(\boldsymbol{e}\left(w_{k}\right)^{\intercal} \boldsymbol{h}^{\text{L}}_{i}\right)}{\sum_{w^{\prime}} \exp \left(\boldsymbol{e}\left(w^{\prime}\right)^{\intercal} \boldsymbol{h}^{\text{L}}_{i}\right)},
\end{equation}
where $\boldsymbol{e}(\cdot)$ denotes the token embedding. The new sentence, with
the inconsistent tokens, is $\boldsymbol{\bar{w}} = \left[\bar{w}_{1}, \ldots, \bar{w}_{n}\right]$:
\begin{equation}
    \label{eq:simneg}
    \bar{w}_{i} \sim p_{\mathrm{MLM}}\left(w \mid \boldsymbol{h}^{\text{L}}_{i}\right).
\end{equation}
The positions of the inconsistent tokens $\mathcal{T}$ are obtained by: 
\begin{equation}
    \label{eq:update_T}
    \mathcal{T} = \left\{t \mid \bar{w}_{t} \neq w_{t}, \forall t \in \mathcal{M} \right\}.
\end{equation}

We also train the language model to fit to the sentence during pre-training by a MLM objective:
\begin{equation}
    \mathcal{L}_{\mathrm{GEN}}=-\sum_{i \in \mathcal{M}} \log
	 p_{\mathrm{MLM}}\left(w_{i} \mid \boldsymbol{h}^{\text{L}}_{i} \right).
    \label{eq:lgen}
\end{equation}

We provide justification for using a language model to approximately fulfill the two conditions of inconsistent tokens.
First, the language model has no access to the visual inputs. $P_{\mathcal{W,
V}}(\boldsymbol{W} = \boldsymbol{\bar{w}}, \boldsymbol{V} = \boldsymbol{v})$ is
likely to be small, and so it generates samples approximating the first condition.
Second, since the language model is fine-tuned during pre-training,
$P_{\mathcal{W}}(\boldsymbol{W} = \boldsymbol{\bar{w}})$ is likely to be large.
Thus, it generates samples that can approximate the second condition.

\subsection{Saliency-based Masking}
\label{saliency}
In this section, we discuss how to obtain masking positions $\mathcal{M}$.
First, we introduce the concept of saliency.

\noindent\textbf{Saliency of Text Tokens w.r.t. Images.}
For a pair of sentence $\boldsymbol{w}$ and image $\boldsymbol{v}$, a text token
$w_{i}$ is salient w.r.t. to the image if the meaning of the token is strongly related to the content in the image, otherwise, it is not salient.
In Fig. \ref{fig:overview}, words ``yellow'', ``blue'', ``hydrant'' and ``grass''
are salient w.r.t. the image, while ``and'', ``the'' are not.

\noindent\textbf{Masking Salient Tokens.}
As discussed in Sec. \ref{negative}, we first mask the original sentence based on
$\mathcal{M}$, and obtain the positions $\mathcal{T}$ for inconsistent tokens by
comparing the generated tokens with the original tokens (Eq. \eqref{eq:update_T}).
Since the language model is fine-tuned on the corpus, if we mask non-salient
tokens w.r.t. an image, the language model is likely to recover the original
tokens by attending to the remaining language context.
In this case, no inconsistent tokens will be generated, and the model cannot learn vision-language associations.
However, if we mask salient tokens w.r.t. an image, the language model is incapable of recovering such tokens because it has no access to the visual input. 
Therefore, it is expected to mask salient tokens for generating inconsistent tokens.

In practice, we find salient tokens by selecting the tokens/positions with higher
attention scores to the image.
The cross-modal representation of the \texttt{[CLS]} token of the image
$\boldsymbol{v}$ is query $\boldsymbol{q}^{V}_{k} \in \mathbb{R}^{d_{h}}$, and that of all the elements in $ \boldsymbol{w}$ are keys $\boldsymbol{K}^{W}_{k} \in \mathbb{R}^{n\times d_{h}}$ at head $k$, where $n$ is the sequence length of the sentence, $d_h = d/h$ is the dimension of a single-head output.
For simplicity, we only consider the representations from the penultimate layer of the fusion module.
The image-token saliency $\boldsymbol{\alpha} \in \mathbb{R}^{d}$ can then be written as
\begin{equation}
    \label{eq:alpha}
    \boldsymbol{\alpha} = \operatorname{softmax}\left(\frac{1}{h}\sum_{k=1}^{h} \boldsymbol{q}^{V}_{k}(\boldsymbol{K}^{W}_{k})^{\intercal}\right).
\end{equation}

Finally, the masking positions $\mathcal{M}$ are sampled without replacement from the categorical distribution $p(t = i; \boldsymbol{\alpha}) = \alpha_{i}$ such that $|\mathcal{M}| = m$, where $m$ is the expected number of inconsistent tokens.
In this way, tokens with high saliencies are more likely to be masked.
Notice that we use a pre-trained vision-language model to obtain $\boldsymbol{q}^{V}_{k}$ and $\boldsymbol{K}^{W}_{k}$. 
Details are in 
Appendix \ref{apx:tcher}.

\section{Experiments}
\label{sec:exp}
\subsection{Pre-training Settings}
\label{exp:setting}
\noindent\textbf{Baselines.}
The proposed method has no constraints on the network architectures, training objectives, and visual representations. 
It supports architectures conducting either cross-attention or self-attention for multi-modal fusion. 
In this study, we plug-in our method into four recent approaches with diversified
cross-modal learning techniques: (i) METER, \cite{dou2022meter} which adopts
modality-specific encoders with cross-attention, (ii) ALBEF \cite{Li2021AlignBF},
which utilizes cross-modal contrastive learning with cross-attention, (iii) X-VLM
\cite{xvlm},
which performs cross-modal alignments from multiple granularities with
cross-attention, and (iv) ViLT \cite{kim2021vilt}, which fuses images and text in a single encoder via self-attention.
We reproduce the pre-training for all the baselines with the settings where they achieve their best results.

\begin{table*}[!htpb]
    \small
	\centering	
	\resizebox{\textwidth}{!}{
    	\begin{tabular}	{l l c |  c  c | c  c | c c }
    		\toprule	 	
        	 \multirow{3}{*}{Method} & \multirow{3}{*}{Data} & \multirow{3}{*}{EPIC} & \multicolumn{2}{c|}{MSCOCO (5K test set)} & \multicolumn{2}{c|}{Flickr30K (1K test set)} & \multicolumn{2}{c}{Flickr30K ZS(1K test set)} \\
        	 & & &  TR & IR & TR & IR & TR & IR\\
        	 & & & R@1/R@5/R@10 & R@1/R@5/R@10 & R@1/R@5/R@10 & R@1/R@5/R@10 & R@1/R@5/R@10 & R@1/R@5/R@10\\
        	 \midrule
        	
        	\multirow{4}{*}{METER}   
        	& \multirow{2}{*}{4M}   & \xmark & 77.2 / 93.7 / 97.1 & 59.2 / 84.0 / 90.8 & 94.2 / \textbf{99.6} / \textbf{99.9} & 83.8 / 97.3 / 98.6  & 92.2 / \textbf{99.3} / 99.7 & 78.8 / \textbf{94.5} / 96.9 \\
        	&                       & \cmark & \textbf{79.0} / \textbf{94.5} / \textbf{97.5} & \textbf{61.2} / \textbf{85.2} / \textbf{91.6} & \textbf{95.8} / 99.3 / 99.6 & \textbf{85.1} / \textbf{97.4} / \textbf{98.7}  & \textbf{93.1} / 99.1 / \textbf{99.8} & \textbf{79.0} / \textbf{94.5} / \textbf{97.1} \\
        	& \multirow{2}{*}{16M}  & \xmark & 78.2 / 93.8 / 96.9 & 59.8 / 84.2 / 90.8 & 94.6 / 99.7 / 99.8 & 85.2 / 97.4 / 98.8  & 93.1 / 99.4 / \textbf{99.8} & 81.1 / \textbf{96.0} / 97.8 \\
        	&                       & \cmark & \textbf{79.7} / \textbf{94.8} / \textbf{97.5} & \textbf{62.5} / \textbf{85.4} / \textbf{91.9} & \textbf{96.5} / \textbf{99.9} / \textbf{99.9} & \textbf{86.7} / \textbf{97.8} / \textbf{99.0}  & \textbf{94.7} / \textbf{99.5} / \textbf{99.8} & \textbf{82.6} / 95.9 / \textbf{99.8} \\
        	\midrule
        	
        	\multirow{4}{*}{ALBEF} 
        	& \multirow{2}{*}{4M}   & \xmark & 73.3 / 92.4 / \textbf{96.4} & 56.4 / 81.8 / 88.9 & 94.4 / 99.4 / 99.8 & 82.1 / 95.7 / 97.9  & 91.1 / 98.7 / 99.4 & 76.1 / 93.0 / 95.9 \\
        	&                       & \cmark & \textbf{75.1} / \textbf{92.9} / \textbf{96.4} & \textbf{58.6} / \textbf{82.7} / \textbf{89.3} & \textbf{95.6} / \textbf{99.7} / \textbf{99.9} & \textbf{83.7} / \textbf{96.7} / \textbf{98.4}  & \textbf{91.7} / \textbf{99.2} / \textbf{99.8} & \textbf{78.1} / \textbf{93.8} / \textbf{96.4} \\
        	& \multirow{2}{*}{16M}  & \xmark & 78.3 / 93.9 / 96.8 & 61.3 / 84.3 / 90.6 & 95.9 / 99.8 / \textbf{100} & 85.7 / 97.2 / 98.8  & 93.6 / 99.5 / 99.8 & 83.2 / 95.9 / 97.7 \\
        	&                       & \cmark & \textbf{79.2} / \textbf{94.7} / \textbf{97.5} & \textbf{62.9} / \textbf{85.4} / \textbf{91.3} & \textbf{96.4} / \textbf{100} / \textbf{100} & \textbf{87.1} / \textbf{97.3} / \textbf{98.9}  & \textbf{94.8} / \textbf{99.7} / \textbf{99.9} & \textbf{84.1} / \textbf{96.5} / \textbf{97.9} \\
        	\midrule

        	\multirow{4}{*}{X-VLM} 
        	& \multirow{2}{*}{$\text{4M}^{+}$}   & \xmark & 79.8 / 95.1 / 97.7 & 62.7 / 85.6 / 91.4 & 96.7 / \textbf{99.9} / \textbf{100} & 85.3 / 97.4 / \textbf{98.7}  & 83.1 / 97.8 / 99.4 & 70.5 / 92.8 / 96.4 \\
        	&                       & \cmark & \textbf{81.0} / \textbf{95.3} / \textbf{97.9} & \textbf{64.1} / \textbf{86.1} / \textbf{91.6} & \textbf{97.2} / \textbf{99.9} / \textbf{100} & \textbf{87.0} / \textbf{97.6} / \textbf{98.7}  & \textbf{86.2} / \textbf{98.6} / \textbf{99.8} & \textbf{74.4} / \textbf{94.3} / \textbf{97.0} \\
        	& \multirow{2}{*}{$\text{16M}^{+}$}  & \xmark & 79.5 / 95.4 / 97.8 & 63.3 / 85.6 / 91.4 & 96.8 / \textbf{100} / \textbf{100} & 86.7 / 97.5 / 98.7  & 86.4 / \textbf{99.2} / 99.6 & \textbf{76.1} / 94.1 / 96.7 \\
        	&                       & \cmark & \textbf{80.7} / \textbf{95.6} / \textbf{98.0} & \textbf{64.1} / \textbf{85.9} / \textbf{91.8} & \textbf{97.4} / \textbf{100} / \textbf{100} & \textbf{87.3} / \textbf{97.6} / \textbf{98.8}  & \textbf{89.0} / 99.0 / \textbf{99.7} & 75.4 / \textbf{94.2} / \textbf{96.8} \\
        	\midrule
        	
        	\multirow{2}{*}{ViLT} 
        	& \multirow{2}{*}{4M}   & \xmark & 60.4 / 85.8 / 92.2 & 41.3 / 71.4 / 82.3 & 80.8 / 95.9 / 98.7 & 61.2 / 88.0 / 93.5  & 72.6 / 93.0 / 96.8 & 53.4 / 80.8 / 88.7  \\
        	&                       & \cmark & \textbf{65.0} / \textbf{87.5} / \textbf{93.7} & \textbf{46.0} / \textbf{74.8} / \textbf{84.6} & \textbf{85.2} / \textbf{97.3} / \textbf{99.3} & \textbf{66.9} / \textbf{90.0} / \textbf{94.4}  & \textbf{79.8} / \textbf{95.8} / \textbf{97.9} & \textbf{63.3} / \textbf{86.3} / \textbf{92.0} \\
        	
    		\bottomrule
    	\end{tabular}
	}
	\caption
	{Image-text retrieval results on the MSCOCO (fine-tuned) and Flickr30K (fine-tuned and zero-shot) datasets. IR: Image Retrieval and TR: Text Retrieval. Recall@$K$ with $K$ = 1, 5, and 10 is used as the evaluation metric. Better results under the same baseline are marked in \textbf{bold}.
	}
	\label{tbl:ftirtr}
\end{table*}

\noindent\textbf{Datasets.}
There are four widely adopted datasets for vision-language pre-training: (i) COCO
\cite{lin2014microsoft}, (ii) Visual Genome (VG) \cite{krishna2017visual}, (iii)
SBU Captions \cite{ordonez2011im2text} and (iv) Conceptual Captions 3M (CC) \cite{changpinyo2021conceptual}.
We refer the combination of these datasets as 4M because there are 4 million unique images in them. 
We also experiment
with a large-scale web dataset CC12M \cite{changpinyo2021conceptual}.
We consider the combination of 4M and 12M datasets as 16M.
Additionally, the X-VLM requires fine-grained annotations (e.g., object and region
descriptions) for both the 4M and 16M settings.
Therefore, the resulting dataset for X-VLM is called $\text{4M}^{+}$ and
$\text{16M}^{+}$, respectively. 
A more detailed description of the data statistics is in Appendix \ref{apx:data}.

\noindent\textbf{Implementation Details.}
For the pre-training baselines, we follow the official implementations provided by the authors of METER, X-VLM, ALBEF and ViLT. 
More details on their implementations are in Appendix \ref{apx:baseline}.
For the proposed method, the auxiliary language model is chosen to be identical to the text encoder of the vision-language model.
For ViLT without modality-specific encoders, we use BERT-base \cite{devlin2018bert}.
All the auxiliary language models are loaded directly from HuggingFace repository \cite{wolf-etal-2020-transformers} with pre-trained checkpoints.
For every sentence, the number of masked tokens $m$ is calculated as $m = \operatorname{ceil}( \text{mask\_ratio} \times \text{len(sentence))}$. 
The mask ratio is set to $0.35$ for all baselines (based on the 
hyper-parameter search
results
in Sec. \ref{exp:abla}).
The loss weight $\lambda$ of the Image-Token Consistency task is set to $8$.

\begin{table}[!t]
    \small
	\centering	
 \setlength\tabcolsep{3pt}
	\resizebox{1.0\columnwidth}{!}{%
	\begin{tabular}	{l  l c |  c  c  c  c  c  c  }
		\toprule	 	
	 \multirow{3}{*}{Method} & \multirow{3}{*}{Data} & \multirow{3}{*}{EPIC} & \multicolumn{6}{c}{MSCOCO (5K test set)}\\
	 & & &  \multicolumn{3}{c}{TR}& \multicolumn{3}{c}{IR} \\
	 & & & R@1 &R@5&R@10& R@1 &R@5&R@10 \\
	 \midrule
	\multirow{4}{*}{X-VLM} & \multirow{2}{*}{$\text{4M}^{+}$} & \xmark & 69.2 & 91.9 & 96.5 & 55.3 & 82.5 & 89.7 \\
	& & \cmark & \textbf{72.0} & \textbf{93.4} & \textbf{97.3} & \textbf{57.3} & \textbf{83.4} & \textbf{90.2} \\
	& \multirow{2}{*}{$\text{16M}^{+}$} & \xmark & 73.1 & 92.8 & \textbf{97.0} & 56.9 & 83.0 & 89.9 \\
	& & \cmark & \textbf{73.2} & \textbf{93.6} & 96.9 & \textbf{57.5} & \textbf{83.3} & \textbf{90.0} \\
	\midrule
	\multirow{2}{*}{ViLT} & \multirow{2}{*}{4M} & \xmark & 54.6 & 81.4 & 89.1 & 38.6 & 69.0 & 80.2 \\
	& & \cmark & \textbf{64.1} & \textbf{86.5} & \textbf{92.5} & \textbf{47.1} & \textbf{74.8} & \textbf{84.6} \\
	\bottomrule
	\end{tabular}
 	}
 	 \vspace{-2ex}
	\caption
	{
	\small	
		Zero-shot image-text retrieval results on MSCOCO.
	}
	\label{tbl:retrieval_zs}
	\vspace{-10pt}
\end{table}		

\subsection{Results on Downstream Tasks}
Evaluation is performed on the following downstream tasks: (i) Image-Text
Retrieval, (ii) Visual Question Answering (VQA), (iii) Natural Language for Visual
Reasoning (NLVR2), and (iv) Visual Entailment (VE).
Details on these tasks are in Appendix \ref{apx:tasks}.

\subsubsection{Results for Image-Text Retrieval}
Tables \ref{tbl:ftirtr} and \ref{tbl:retrieval_zs} show that EPIC is universally
effective over the different baselines/datasets, since we achieve non-trivial improvement nearly for all the settings.
In addition, compared with all the baselines, EPIC brings more significant improvement to ViLT on different datasets. 
For example, on the fine-tuned retrieval tasks, we achieve an absolute improvement of around 5\% in terms of the TR/IR @1 metric on MSCOCO and Flickr30K.
Furthermore, for zero-shot tasks, the improvement 
on the Flickr30K dataset
is 7.2\% in terms of TR@1, and 9.9\% in terms of IR@1.
The gap between ViLT and the other baselines can be explained by the fact that ViLT adopts a simplified architecture for vision-language pre-training 
and it does not incorporate more sophisticated tasks to learn vision-language associations.
Nevertheless, the results on ViLT can be seen as an indicator for the effectiveness of the proposed method on a ``clean'' baseline.

For the other baselines with advanced strategies to learn vision-language
associations (such as cross-modal contrasting in ALBEF, fine-grained reasoning in
X-VLM, and powerful pre-trained encoders in METER), EPIC still demonstrates
significant improvements.
For example, compared with METER on the MSCOCO dataset (fine-tuned), EPIC achieves an absolute 
IR@1 
improvement 
of 2\% and 2.7\% 
with 4M and 16M data, respectively.
When using the X-VLM as baseline, we observe an absolute improvement of 3.9\% and
3.1\% in terms of IR@1 and TR@1, respectively, on Flickr30K (zero-shot).

\subsubsection{Results for VQA, NLVR2, and VE}
Table \ref{tbl:vltask} shows that EPIC is effective among all the vision-language tasks.
We improve over METER in terms of the VQA dev accuracy by 0.3\% under the 16M setting and X-VLM by 0.4\% under the $\text{4M}^{+}$ setting. 
Such improvement is non-trivial given the fact that these two baselines are quite competitive on this task.
We also notice that EPIC improves ALBEF on VQA by 0.9\% in dev accuracy. 
Further, we observe significant improvements over all baselines on NLVR2
(e.g., +2.3\% dev accuracy on METER 16M; +2.2\% std accuracy on ALBEF 4M).
The proposed method is also effective on the SNLI-VE dataset. It brings an absolute improvement of 0.8\% over ALBEF under the 16M setting.  

\begin{table}[!t]
    \small
	\centering	
 \setlength\tabcolsep{3pt}
	\begin{tabular}	{l l c |  c  c  c  c  c  c  }
		\toprule	 	
	 \multirow{2}{*}{Method} & \multirow{2}{*}{Data} & \multirow{2}{*}{EPIC} & \multicolumn{2}{c}{VQA} & \multicolumn{2}{c}{NLVR2} & \multicolumn{2}{c}{SNLI-VE}\\
	 & & & dev & std & dev & std & dev & std   \\
	 \midrule
	\multirow{4}{*}{METER} & \multirow{2}{*}{$\text{4M}^{+}$} & \xmark & 77.7 & 77.9 & 81.8 & 82.5 & 81.4 & 81.0 \\
	& & \cmark & \textbf{77.9} & \textbf{78.0} & \textbf{83.5} & \textbf{83.5} & \textbf{81.6} & \textbf{81.8} \\
	& \multirow{2}{*}{$\text{16M}^{+}$} & \xmark & 78.3 & 78.4 & 82.7 & 84.3 & 81.7 & 81.8 \\
	& & \cmark & \textbf{78.6} & \textbf{78.7} & \textbf{85.0} & \textbf{85.2} & \textbf{82.1} & \textbf{82.3} \\
	\midrule
	\multirow{4}{*}{ALBEF} & \multirow{2}{*}{$\text{4M}^{+}$} & \xmark & 74.6 & 74.6 & 79.5 & 80.0 & 80.1 & 80.1 \\
	& & \cmark & \textbf{75.1} & \textbf{75.2} & \textbf{81.3} & \textbf{82.2} & \textbf{80.6} & \textbf{80.7} \\
	& \multirow{2}{*}{$\text{16M}^{+}$} & \xmark & 75.8 & 76.0 & 82.6 & 82.5 & 80.8 & 80.9 \\
	& & \cmark & \textbf{76.7} & \textbf{76.7} & \textbf{84.1} & \textbf{84.0} & \textbf{81.3} & \textbf{81.7} \\
	\midrule
	\multirow{4}{*}{X-VLM} & \multirow{2}{*}{$\text{4M}^{+}$} & \xmark & 78.1 & 78.2 & 83.3 & 84.1 & / & / \\
	& & \cmark & \textbf{78.5} & \textbf{78.5} & \textbf{84.6} & \textbf{84.5} & / & / \\
	& \multirow{2}{*}{$\text{16M}^{+}$} & \xmark & 78.0 & 78.2 & 84.3 & 84.5 & / & / \\
	& & \cmark & \textbf{78.3} & \textbf{78.3} & \textbf{85.2} & \textbf{85.5} & / & / \\
	\midrule
	\multirow{2}{*}{ViLT} & \multirow{2}{*}{4M} & \xmark & 71.3 & 71.4 & 75.0 & 75.2 & / & / \\
	& & \cmark & \textbf{71.8} & \textbf{71.8} & \textbf{77.2} & \textbf{77.1} & / & / \\
	\bottomrule
	\end{tabular}
 	 \vspace{-2ex}
	\caption
	{
	\small	
		Evaluation results on downstream vision-language tasks: VQA, NLVR2, and SNLI-VE. ``/'' indicates that the original baseline does not conduct experiment on this task.
	}
	\label{tbl:vltask}
\end{table}	

\begin{table}[!h]
    \small
	\centering	
 \setlength\tabcolsep{3pt}
	\resizebox{1.0\columnwidth}{!}{%
	\begin{tabular} {l | c  c  c  c  c  c  c  }
		\toprule	 	
	 \multirow{2}{*}{Method} & NLVR2 & \multicolumn{2}{c}{\small{Flickr30K-ft}} & \multicolumn{2}{c}{\small{Flickr30K-zs}} & \multicolumn{2}{c}{\small{MSCOCO-ft}}\\
	 & dev & TR1 & IR1 & TR1 & IR1 & TR1 & IR1 \\
	 \midrule
	\small{vanilla METER} & 79.6 & 89.2 & 76.6 & 83.2 & 67.7 & 71.0 & 52.5 \\
	\small{ITC (rand.)} & 79.9 & 91.5 & 77.8 & 83.2 & 69.0 & 70.8 & 53.5 \\
	\small{ITC+LM} & 80.9 & 92.1 & 78.9 & 83.8 & 71.5 & 73.4 & \textbf{55.6} \\
	\midrule
	\small{EPIC} & \textbf{81.0} & \textbf{92.9} & \textbf{79.0} & \textbf{84.3} & \textbf{72.5} & \textbf{74.1} & \textbf{55.6} \\
	\bottomrule
	\end{tabular}
 	}
 	 \vspace{-2ex}
	\caption
	{
	\small	
		Ablation studies on the effect of the ITC task, negative samples generation and the saliency-based masking strategy.
	}
	\label{tbl:abl_comp}
\vspace{-10pt}
\end{table}		

\begin{table}[!t]
    \small
	\centering	
 \setlength\tabcolsep{3pt}
	\resizebox{1.0\columnwidth}{!}{%
	\begin{tabular} {l | c  c  c  c  c  c  c  }
		\toprule	 	
	 \multirow{2}{*}{Generator} & NLVR2 & \multicolumn{2}{c}{\small{Flickr30K-ft}} & \multicolumn{2}{c}{\small{Flickr30K-zs}} & \multicolumn{2}{c}{\small{MSCOCO-ft}}\\
	 & dev & TR1 & IR1 & TR1 & IR1 & TR1 & IR1 \\
	 \midrule
	\small{LM (cond.)} & 79.9 & 92.4 & 78.7 & 84.0 & 71.8 & 72.5 & 54.5 \\
	\small{VLM (SAS)} & 80.5 & 91.0 & 79.2 & 83.2 & 70.4 & 72.4 & 54.9 \\
	\small{LM (trained)} & 80.7 & 91.3 & \textbf{79.5} & \textbf{84.8} & 72.1 & 73.6 & 55.1 \\
	\small{LM (fixed)} & 80.7 & 91.9 & 79.0 & 83.8 & 71.1 & 72.4 & 54.8 \\
	\midrule
	\small{LM (fine-tune)} & \textbf{81.0} & \textbf{92.9} & 79.0 & 84.3 & \textbf{72.5} & \textbf{74.1} & \textbf{55.6} \\
	\bottomrule
	\end{tabular}
 	}
 	 \vspace{-2ex}
	\caption
	{
	\small	
		Ablation study on different negative sample generators.
	}
	\label{tbl:abl_gen}
\end{table}			

\begin{figure*}[!]
  \centering
  \begin{minipage}{.48\linewidth}
    \centering
    
    \subcaptionbox{\label{fig:train}}
      {\includegraphics[width=0.28\linewidth, height=0.30\linewidth]{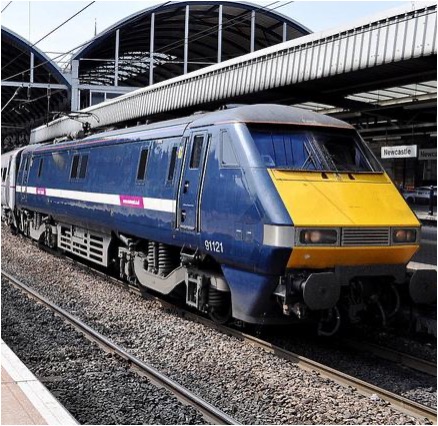}}
    \hfill
      {\includegraphics[width=0.7\linewidth, height=0.30\linewidth]{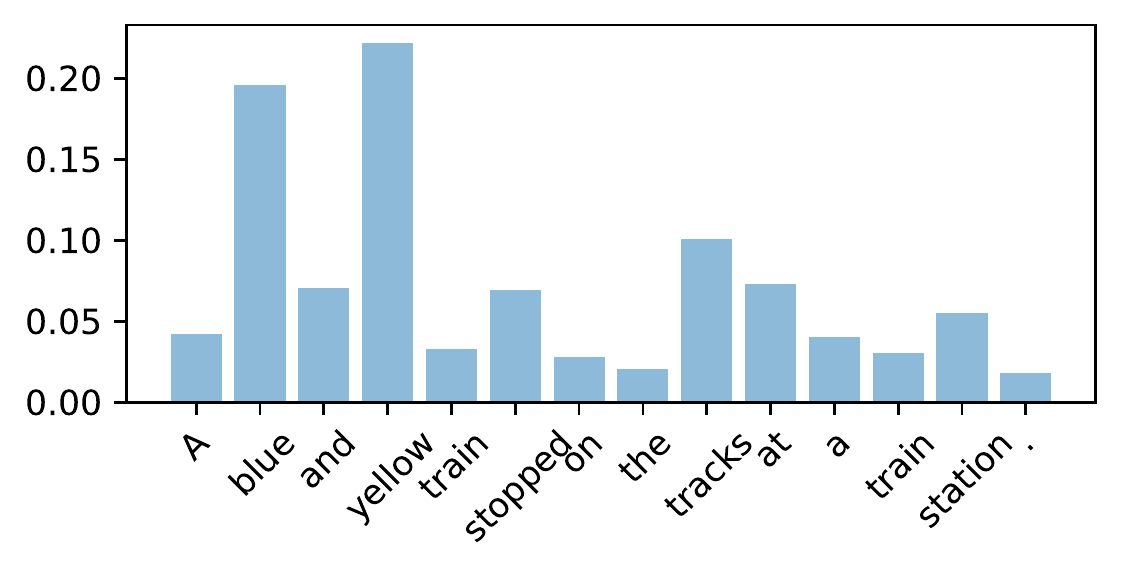}}
     
    \subcaptionbox{\label{fig:man}}
      {\includegraphics[width=0.28\linewidth, height=0.30\linewidth]{coco_man_mosaic.png}}
    \hfill
      {\includegraphics[width=0.7\linewidth, height=0.30\linewidth]{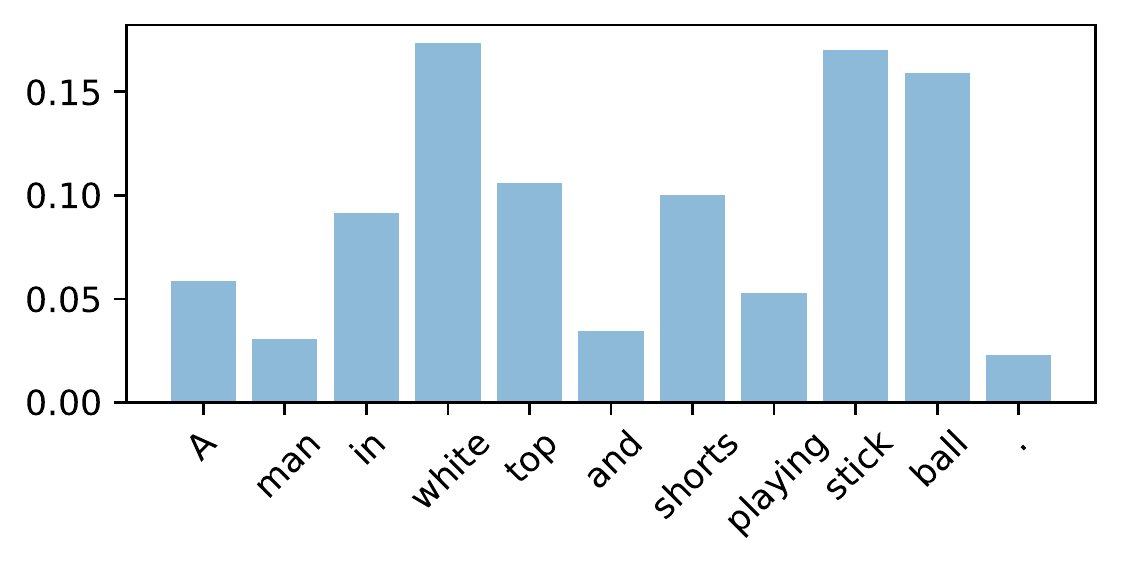}}


    \caption{Visualization on the token saliency distribution.}
    \label{fig_visual_saliency}
  \end{minipage}\quad
    \vspace{-5pt}
  \begin{minipage}{.48\linewidth}
    \centering
    

      {\includegraphics[width=\linewidth,height=0.34\linewidth]{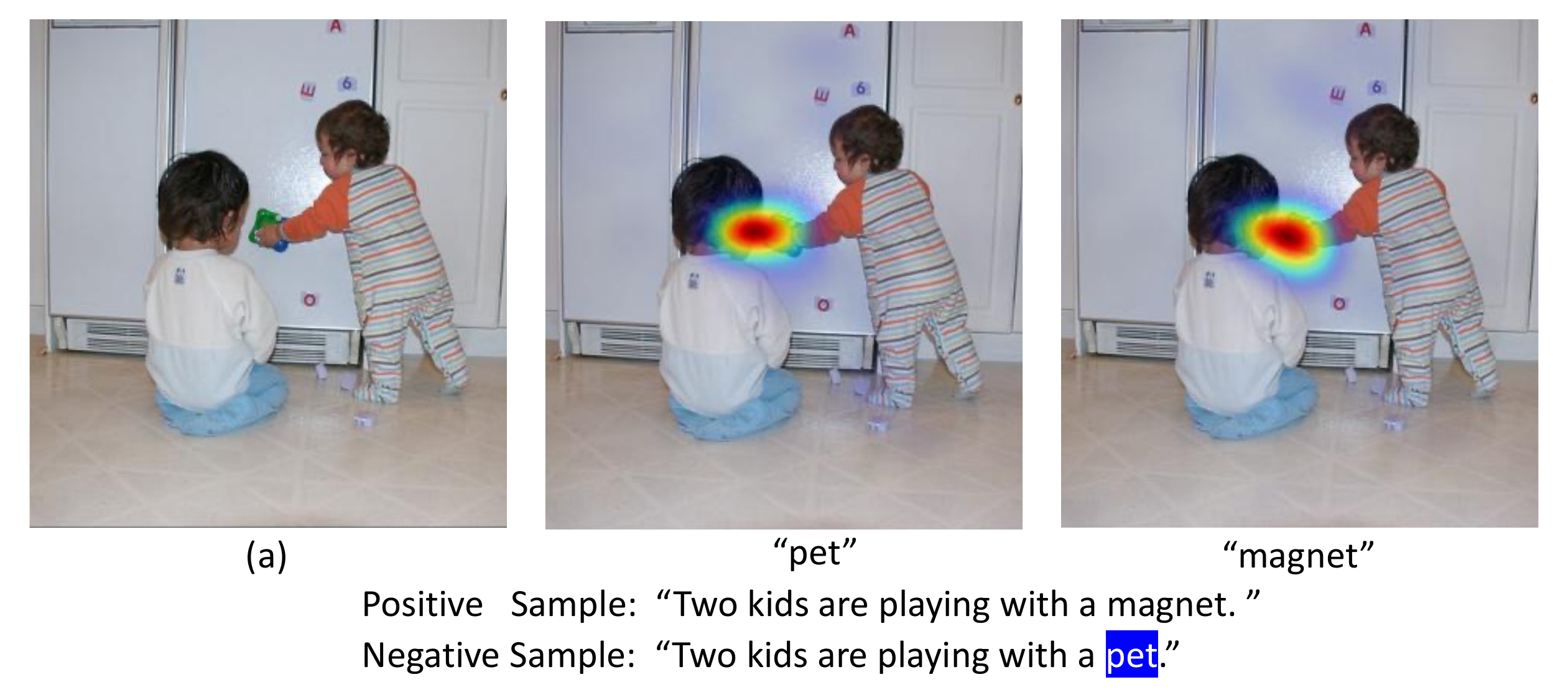}}

      {\includegraphics[width=\linewidth,height=0.36\linewidth]{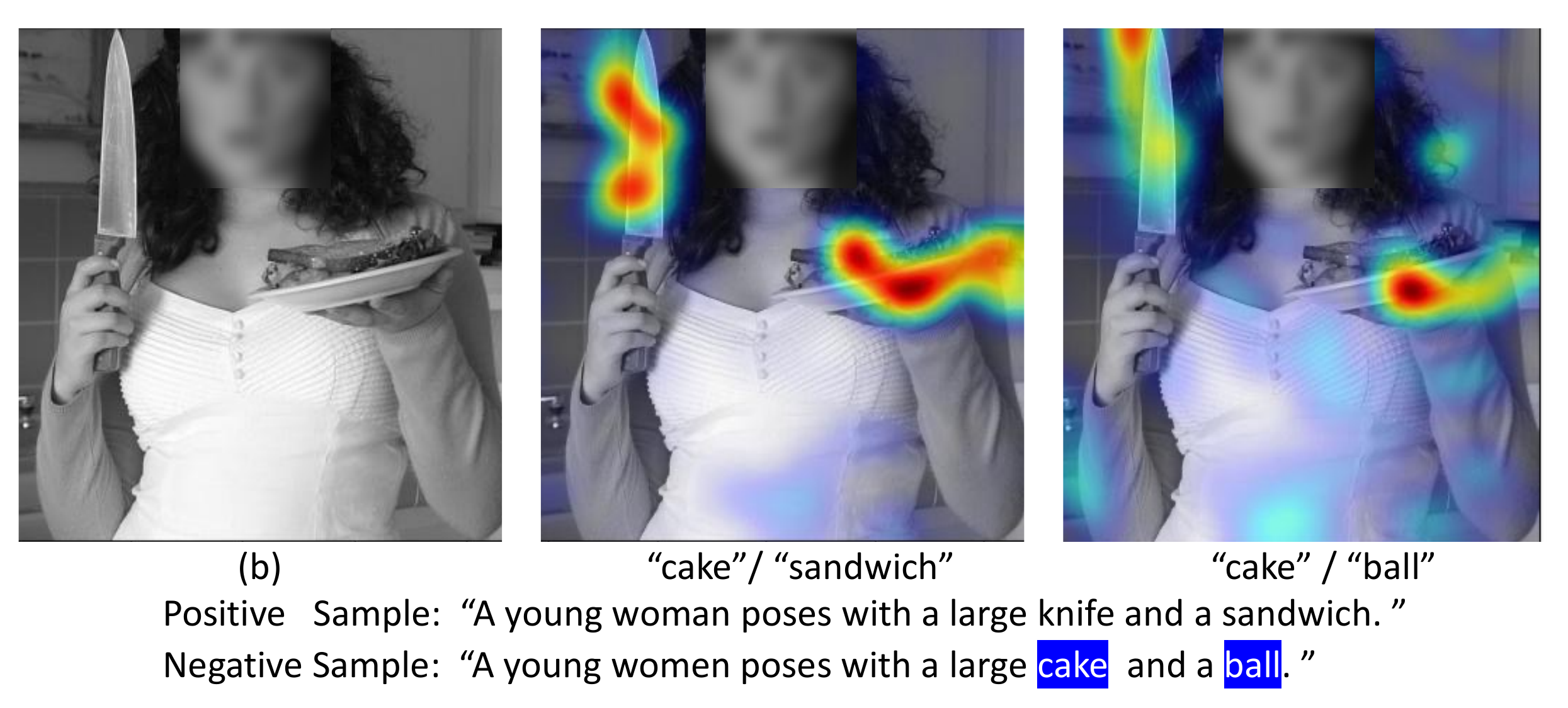}}
    
    \caption{Grad-CAM visualization on the cross-attention maps of the image with respect to the ITC task.}
    \label{fig:gradcam}
  \end{minipage}
  \vspace{-5pt}
\end{figure*}

\subsection{Ablation Study}
\label{exp:abla}

We conduct ablation studies based on the METER model due to its superior performance with basic training objectives (i.e., image-text matching and cross-modal masked language modeling). 
However, training a full-version of METER is expensive (roughly 4 days
with 32 A100 80G). Hence, we conduct ablations on a smaller scale that is still representative.
Specifically, the model is still pre-trained on the 4M dataset, but the input
image is resized to $224 \times 224$ (instead of $288 \times 288$
in the normal METER setting).
To speed up the training process,
we replace the image encoder CLIP-16 \cite{radford2021learning} with CLIP-32 for
less memory usage, and also shorten the training schedule to 50k steps.

We evaluate the pre-trained model on the NLVR2 task and retrieval task (Flickr30K fine-tuned/zero-shot, MSCOCO fine-tuned).
For the NLVR2 task, we report the development accuracy. For the retrieval tasks,
we report the IR1/TR1 accuracies on the corresponding validation sets.

\noindent\textbf{Component Analysis.}
As shown in Table \ref{tbl:abl_comp}, on top of vanilla
METER, we first add an ITC task 
(\emph{ITC (rand.)})
with inconsistent tokens sampled uniformly from the vocabulary (selecting 35\% of positions for inconsistent tokens).
In this case, though some of the sampled tokens do not satisfy the conditions for
inconsistent tokens in Sec.~\ref{itc}, the pre-trained model can still benefit from the ITC task.
When we replace the random strategy with a language model, \emph{ITC+LM}, we generate inconsistent tokens approximating the conditions in Sec.~\ref{itc}.
Finally, we choose to mask the tokens that are salient w.r.t. the image to generate inconsistent tokens, \emph{ITC+saliency+LM}, and this further improves the performance on downstream tasks.
Therefore, each component of EPIC is effective to improve the performance of the pre-trained model.

\noindent\textbf{Conditions of Inconsistent Tokens.}
In this ablation experiment, we study the importance of the two conditions of inconsistent tokens in Sec.~\ref{itc}.
\emph{LM (fine-tune)} is proposed in EPIC and it produces inconsistent tokens approximating the two conditions.
First, we drop the first condition, that is, $P_{\mathcal{W, V}}(\boldsymbol{W} = \boldsymbol{\bar{w}}, \boldsymbol{V} = \boldsymbol{v})$ is no longer small.
We achieve this in two ways. (i)
\emph{LM (cond.)}: When generating inconsistent samples and fine-tuning the LM, we replace the class token (first token) in the text with the one from the image encoder of the VLM (with gradient propagation canceled).
(ii) \emph{VLM (SAS)}: We adapt SAS \cite{xu2021sas} from the text modality to the multi-modality setting.
An auxiliary VLM (previous checkpoints of the VLM, detailed in Appendix \ref{apx:sas}) is used to produce inconsistent tokens.
As shown in Table \ref{tbl:abl_gen}, we observe that both \emph{LM (cond.)} and
\emph{VLM (SAS)} suffer performance deterioration compared to \emph{LM (fine-tune)}.
This indicates the importance of the first condition.

We then demonstrate the importance of the second condition, that is $P_{\mathcal{W}}(\boldsymbol{W} = \boldsymbol{\bar{w}})$ is large.
Intuitively, we fail to satisfy this condition when we stop fine-tuning the LM on the text corpus, i.e., \emph{LM (fixed)}.
Further, for \emph{LM (fine-tune)}, we experiment its possible alternatives, \emph{LM (trained)},
where we fit the LM on the text corpus before instead of during pre-training.
We observe that \emph{LM (fixed)} clearly decreases the performance (especially in MSCOCO-ft).
This validates the effectiveness of the second condition.
Note that \emph{LM (fine-tune)} outperforms \emph{LM (trained)} by a small margin.
This gap can be attributed to the training dynamics brought by fine-tuning
\cite{meng2021coco}. As the LM fits the text corpus gradually better during
finetuning, it becomes increasingly hard for a model to identify whether a token
is replaced. This dynamics encourages the model to perform curriculum learning, which improves performance.


\subsection{Visualization}
\noindent
\textbf{Token Saliency.}
Fig. \ref{fig_visual_saliency}
shows the saliency distribution of the tokens w.r.t. the image
for an image-sentence pair.
We can see that salient tokens have higher densities in the distribution.
For example, in Fig. \ref{fig:train}, ``blue'' and ``yellow'' are highlighted. In Fig. \ref{fig:man}, the teacher model gives more mass on ``white'', ``stick'' and ``ball''. 
These confirm that salient tokens can be detected by EPIC.

\noindent
\textbf{Grad-CAM of ITC.}
Fig.~\ref{fig:gradcam} visualizes the cross-attention maps of the image w.r.t. the inconsistent/consistent tokens using Grad-CAM \cite{selvaraju2017grad}. 
In each row, we show the original image (left), the attention map w.r.t. the
inconsistent tokens (middle), and that w.r.t. the consistent ones (right). 
For example, in the first row, when we input the model with the original sentence,
the model predicts that the token ``magnet'' is consistent and attends to the actual magnet in the image.
However, when we replace ``magnet'' with ``pet'', the model predicts the token
``pet'' as ``inconsistent'' and still attends to the magnet in the image. This means the model is aware that the object in the image is actually a magnet instead of a pet. Similar observation exists for the second row. 
These demonstrate the model's ability to reason on cross-modal relationship.

\vspace{-3pt}
\section{Conclusion}
In this paper we propose \textbf{EPIC}, a pre-training approach that leverage more
text tokens for learning vision-language associations. It is less affected by the modality bias problem compared with CMLM.
Specifically, we propose an ITC task to identify inconsistent tokens generated by a language model coupled with a saliency-based masking strategy.
The task formulation of the ITC task and the design of inconsistent samples
address the problems of under-utilization of unmasked tokens and modality bias.
We perform extensive experiments and show that EPIC brings consistent performance
gains over several baselines on a wide range of downstream tasks. Possible
directions for future research can be: (i) approaching the conditions of
inconsistent tokens in Sec.~\ref{itc} more precisely; (ii) finding salient tokens without using a pre-trained teacher VLM.

\section*{Acknowledgements}

This work is supported by NSFC key grant 62136005, NSFC general grant 62076118, and Shenzhen fundamental research program JCYJ20210324105000003.

{\small
\bibliographystyle{ieee_fullname}
\bibliography{EPIC}
}

\clearpage

\appendix
\section*{\centering {\Large Appendix}}

\section{Experimental Details for Sec. \ref{sec2}}

For the ``VLM'' mentioned in Sec. \ref{sec2}, we use the architecture designed in METER \cite{dou2022meter}. 
Similar to the ablation study in experiments, the model is pre-trained on the 4M dataset with an image-text matching loss and the (cross-modal) masked language modeling loss, but the input image is resized to $224 \times 224$ instead of $288$ as in the normal setting of METER. 
We replace the image encoder CLIP-16 \cite{radford2021learning} with CLIP-32 for less memory usage and train the model for 100k steps. 
For ``LM'', we use roberta-base directly loaded from HuggingFace \cite{wolf-etal-2020-transformers}. 
When training LM, the corpus come from all the sentences in the 4M dataset by excluding images.
The model is finetuned with the masked language modeling loss. 
Other hyper-parameters are set to identical to that in ``VLM''. 
For evaluation, the accuracy is computed over the validation set of the COCO dataset \cite{lin2014microsoft}.

\noindent\textbf{Identifying Modality Bias.}
We train VLM and LM on the CMLM/MLM task simultaneously, that is, for each mini-batch in the dataloader (including images, sentences, masked tokens), the VLM takes the masked sentences and images but the LM takes only the masked sentences.
The mask probability is set to $15\%$.
We report the per-epoch accuracy.

\noindent\textbf{Under-utilization of Unmasked Tokens.}

This experiment does not involve any training as the pre-trained models of VLM and LM come from Sec.~\ref{sec2}.
The corruption of unmasked tokens are similar to the methods described in Sec.~\ref{negative}. 
Specifically, in the first round, we mask $15\%$ of the tokens for the CMLM/MLM task.
In the second round, we additionally randomly mask $\tau\%$ of the tokens from the tokens that are \emph{NOT} masked in the first round. 
Then we use a pre-trained roberta-base model loaded from \cite{wolf-etal-2020-transformers} to recover the tokens that are masked in the second round.
Some tokens are successfully recovered by roberta-base, while some are not.
The corruption ratio is an average of $\frac{\text{\# tokens are not recovered}}{\text{\# all the tokens in the sentence}}$ among all the sentence in the validation set. 
Finally, this corrupted masked sentence is fed into the VLM/LM for the CMLM/MLM inference.
We record the CMLM/MLM accuracy over corrupted masked sentences and compute the relative performance drop compared to CMLM/MLM accuracy on the clean masked sentence.  
We gradually increase $\tau$ and record the resulting corruption ratio (i.e., x-axis) and relative performance drop (i.e., y-axis). 

\label{apx:emp}

\section{Complete Algorithm of EPIC}
\label{apx:alg}

The algorithm for the EPIC model is shown in Algorithm \ref{algo}, where $\Tilde{f}_{\mathrm{VL}(\cdot)}$ denotes the pre-trained vision-language model to obtain $\boldsymbol{q}^{V}_{k}$ and $\boldsymbol{K}^{W}_{k}$.

\begin{algorithm}[ht]
    \caption{EPIC}\label{algo}
    \textbf{Input}: Token sequence $\boldsymbol{w}$, raw image patches/regions $\boldsymbol{v}$, expected number of inconsistent tokens $m$; \\
    \textbf{Output}: $\mathcal{L}_{\mathrm{ITC}}, \mathcal{L}_{\mathrm{GEN}}$     
    \begin{algorithmic}
        \STATE{// Obtain salient masking positions (Sec. \ref{saliency})}
        \STATE{Obtain $\boldsymbol{q}^{V}_{k}$ and $\boldsymbol{K}^{W}_{k}$ from $\Tilde{f}_{\mathrm{VL}}(\boldsymbol{w}, \boldsymbol{v})$;} 
        \STATE{Compute $\boldsymbol{\alpha}$  based on Eq. \eqref{eq:alpha} and sample $\mathcal{M}$ based on $m$;}

        \STATE{// Generate inconsistent samples (Sec. \ref{negative})}
        \STATE{Obtain $\boldsymbol{H}^{L}$ from $f_{\mathrm{L}}(\boldsymbol{w^{\text{mask}}})$;}
        \STATE{Generate new sentence $\boldsymbol{\bar{w}}$ based on Eqs. \eqref{eq:mlm_prob} and \eqref{eq:simneg};}
        \STATE{Obtain inconsistent tokens positions $\mathcal{T}$ via Eq. \eqref{eq:update_T};}
        \STATE{Compute $\mathcal{L}_{\mathrm{GEN}}$ based on Eq. \eqref{eq:lgen};}
        
        \STATE{// Compute $\mathcal{L}_{\mathrm{ITC}}$ (Sec. \eqref{itc})}
        \STATE{Obtain $\left\{\boldsymbol{\bar{h}}^{\mathrm{VL}}_{i}\right\}_{i=1}^{n}$ from $f_{\mathrm{VL}}(\boldsymbol{\bar{w}}, \boldsymbol{v})$;}
        \STATE{Compute $\mathcal{L}_{\mathrm{ITC}}$ based on Eqs. \eqref{eq:sigmoid} and \eqref{eq:litc};}
    \end{algorithmic}
    \textbf{return $\mathcal{L}_{\mathrm{GEN}}, \mathcal{L}_{\mathrm{ITC}}$ }
\end{algorithm}

\section{Details of Teacher VLM}
\label{apx:tcher}
In our implementation, we use the publicly-available checkpoints of the corresponding baselines as the teacher model. 
The teacher model is frozen and is only leveraged for saliency-based masking during pre-training.

\section{Downstream Tasks}
\label{apx:tasks}
\noindent\textbf{Image-Text Retrieval.}
Image-text retrieval includes two subtasks: (1) retrieving images for given text (Image Retrieval (IR)) and (2) retrieving text for given images (Text Retrieval (TR)).
We conduct two different scenarios for evaluations: “zero-shot” (ZS) retrieval task and “after-finetuning” retrieval task. 
We conduct experiments on the MSCOCO \cite{lin2014microsoft} and Flickr30K \cite{plummer2015flickr30k} datasets.
For evaluation, we use the recall at $K$ (R$@K$) metric, which considers top-$K$ predictions as candidates for correct predictions, and $K$ is chosen from $\{1, 5, 10\}$.
Note that METER, ViLT, and X-VLM directly conduct zero-shot inference with the pre-trained model, while ALBEF uses the model fine-tuned on the MSCOCO dataset for inference.

\noindent\textbf{Visual Question Answering (VQA).}
VQA \cite{antol2015vqa} requires the model to predict an answer given an image and a corresponding question. 
We conduct experiments on the VQA 2.0 dataset \cite{8100153}.
For ALBEF and X-VLM, we treat VQA as a language generation task; for METER and ViLT, we follow the common practice to convert the task into a classification problem. 
The final evaluation scores test-dev (dev) and test-std (std) are obtained from an official evaluation server.

\noindent\textbf{Natural Language for Visual Reasoning (NLVR2).}
This task is to determine whether a natural language caption is true about a pair of photographs.
We use the dataset from \cite{suhr2018corpus}.
Following the practice of baselines, ALBEF and X-VLM conduct further pre-training on the extended pre-trained model and fine-tuned it on the dataset afterwards.
METER and ViLT directly fine-tune the pre-trained model on the dataset.

\noindent\textbf{Visual Entailment (VE).}
Visual entailment is a visual reasoning task to predict whether the relationship between an image and text is entailment, neutral or contradictory.
Of the four baselines, only METER and ViLT conducted fine-tuning on this task, so we follow their procedure to treat this task as a three-way classification problem and report classification accuracy on the SNLI-VE dataset \cite{xie2019visual}.

For all the downstream tasks, we follow original implementations and evaluations of baselines. 

\section{Pre-training Dataset}
In Table \ref{tbl:data}, we provide the statistics of the 4M and 16M datasets.
For details about the $4\text{M}^{+}$ and $16\text{M}^{+}$ dataset, please refer to \cite{xvlm}.

\label{apx:data}

\begin{table}[h]
    \small
	\centering	
	\begin{tabular} {c | c c c}
		\toprule	 	
	 & Dataset & \# Images & \# Captions\\
	 \midrule
	\multirow{4}{*}{4M} & COCO & 0.11M & 0.55M \\
      & VG & 0.10M & 5.7M \\
	  & SBU & 0.86M & 0.86M \\
	  & CC3M & 2.9M & 2.9M \\
	\midrule
	\multirow{2}{*}{16M} & 4M & 4.0M & 10M \\
      & CC12M & 11.1M & 11.M \\
	\bottomrule
	\end{tabular}
	\caption
	{
	\small	
		Statistics of the pre-training dataset.
	}
	\label{tbl:data}
\end{table}		

\section{Implementation Details of Baselines}
\label{apx:baseline}
For, X-VLM and ALBEF, their names refer to the architecture/method proposed in the corresponding paper.
The implementation of X-VLM and ALBEF are identical to the official repository.

\noindent\textbf{METER.}
We reproduced the pre-training of $ \text{METER-CLIP-ViT}_{\mathrm{BASE}} $. 
During pre-training, we additionally apply RandAugment to input images (we remove color changes from RandAugment because the text often contains color information, otherwise we may generate inconsistent tokens unintentionally) as we found that this technique is widely used by a lot of existing vision-language pre-training methods \cite{kim2021vilt, xvlm, Li2021AlignBF} and that it improves the generalization of the pre-trained model.
Further, we found that using a tri-stage learning rate scheduler is beneficial for the performance of downstream tasks. 
Specifically, we linearly warm up the learning rate to its peak value in the first $10\%$ steps, hold this value for the next $80\%$ of the steps, and finally decay it exponentially to 1\% of the peak value in the remaining steps. 

\noindent\textbf{ViLT.}
We tried to reproduce the pre-training of $\text{ViLT-B/32\textcircled{a}\textcircled{+}}$ \cite{kim2021vilt}, but we encountered ``nan`` in the middle of the training process. 
In response, we switched off half-precision (fp16) and continued the training with full-precision (fp32).

Note that the discrepancy in implementations between the original paper and ours does not pose unfairness in comparisons as all the results in Sec. \ref{sec:exp} are obtained based on our reproductions.  

\section{Previous Checkpoints of VLM}
\label{apx:sas}
At the beginning, the auxiliary VLM is the same as the newly-initialized VLM.
After that, at the end of each $k$-th epoch, we save the current model's checkpoint and use it as the auxiliary VLM.
Note that during pre-training, the auxiliary VLM is frozen. 
Therefore, the auxiliary VLM has the same architecture as the main VLM and is able to do the CMLM inference.
Different from the LM as the generators for inconsistent tokens, the auxiliary VLM considers the visual inputs when trying to recover the masked tokens. 

\section{Ablations on different mask ratios}
As shown in Table \ref{tbl:abl_rate}, when the mask ratio increases, more
inconsistent tokens can be generated, which expedites the learning of vision-language associations.
However, when too many tokens are masked, the meaning of the sentence can be completely different.
As a result, the model simply predicts all tokens as inconsistent, harming the pre-training process.
Therefore, the mask ratio is set to $35\%$ as it brings the best (or nearly the
best) results under all settings.

\section{Analysis of ITC Task}

\begin{figure}
     \begin{subfigure}[b]{0.23\textwidth}
         \centering
         \includegraphics[width=\textwidth]{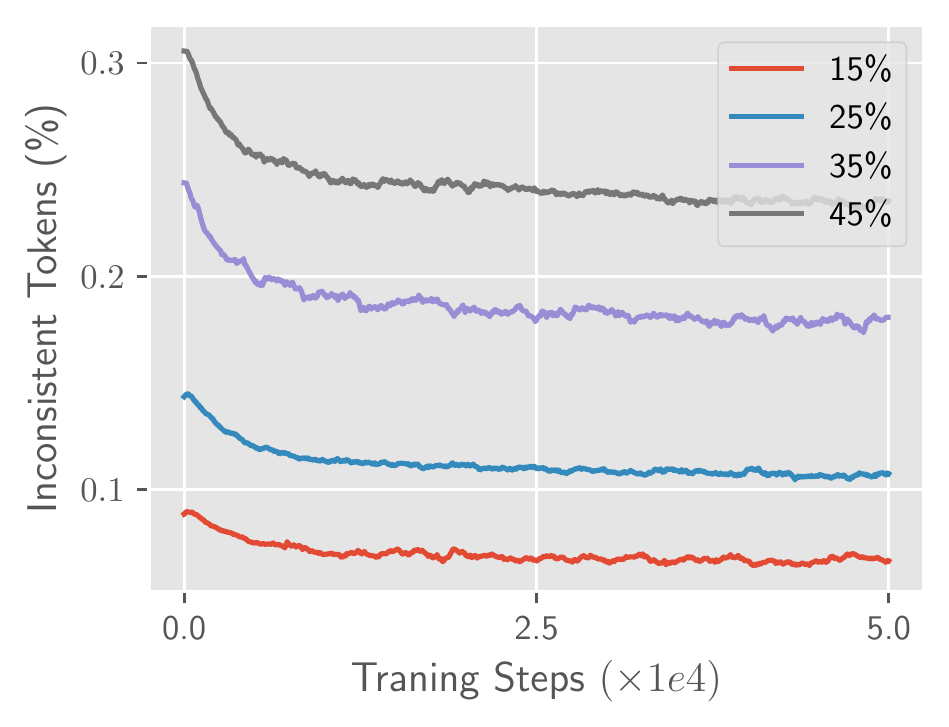}
         \caption{Ratio of Incons. tokens.}
         \label{fig:rate_neg}
     \end{subfigure}
     \hfill
     \begin{subfigure}[b]{0.23\textwidth}
         \centering
         \includegraphics[width=\textwidth]{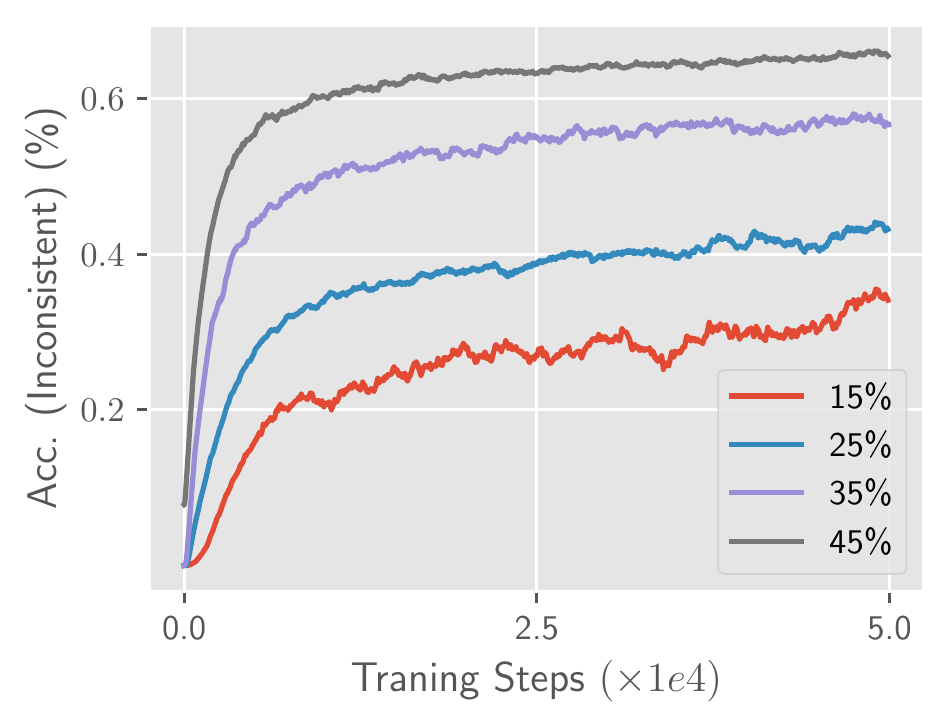}
         \caption{ITC Acc. on Incons. tokens.}
         \label{fig:itc_acc}
     \end{subfigure}
     \hfill
     \caption{Left: The ratio of inconsistent samples (y-axis). Right: the accuracy of the ITC task over the inconsistent samples (y-axis). Colors of the curves indicate different mask rates. }
\end{figure}

Here we empirically analyze the ITC task. In Fig.~\ref{fig:rate_neg}, we show the ratio of inconsistent samples.
This ratio quickly drops in the first few epochs because the language model learns to fit to the text corpus.
However, it converges in later stages as the language model is incapable of recovering the masked salient tokens, thereby producing inconsistent tokens for the ITC task.
Fig.~\ref{fig:itc_acc} shows that the inconsistent tokens are quite challenging for the VLM trained by EPIC to identify.
Of all the inconsistent tokens, only about half (depending on mask rates) could be identified by the VLM because some image-token inconsistency can be subtle.
Such challenge distinguishes ITC from CMLM in that ITC requires sophisticated cross-modal reasoning for each task while only language reasoning could help CMLM to achieve good performance as shown in Fig. \ref{fig:emp_bias_val}.

\begin{table}[!t]
    \small
	\centering	
 \setlength\tabcolsep{3pt}
	\resizebox{1.0\columnwidth}{!}{%
	\begin{tabular} {c | c  c  c  c  c  c  c  }
		\toprule	 	
	 \multirow{2}{*}{Corrupt Rate} & NLVR2 & \multicolumn{2}{c}{\small{Flickr30K-ft}} & \multicolumn{2}{c}{\small{Flickr30K-zs}} & \multicolumn{2}{c}{\small{MSCOCO-ft}}\\
	 & dev & TR1 & IR1 & TR1 & IR1 & TR1 & IR1 \\
	 \midrule
	\small{15\%} & 80.9 & 91.4 & 78.9 & 83.2 & 67.7 & 73.1 & 55.5 \\
	\small{25\%} & \textbf{81.0} & 92.0 & \textbf{79.0} & 83.8 & 71.5 & 73.5 & 55.4 \\
	\small{35\%} & \textbf{81.0} & \textbf{92.9} & \textbf{79.0} & 84.3 & \textbf{72.5} & \textbf{74.1} & 55.5 \\
	\small{45\%} & \textbf{81.0} & 91.6 & \textbf{79.0} & \textbf{84.4} & 71.6 & 73.0 & \textbf{55.8} \\
	\bottomrule
	\end{tabular}
 	}
 	 \vspace{-2ex}
	\caption
	{
	\small	
		Ablations on different mask ratios.
	}
	\label{tbl:abl_rate}
\vspace{-10pt}
\end{table}	

\section{Integration to Baselines}
\label{allloss}
To integrate our method with an existing VLP baseline, one only need to additionally load an auxiliary BERT-like language model and optimize a union of the original objectives and ours. 
For example, in the case of METER \cite{dou2022meter} that is trained by $\mathcal{L}_{\mathrm {ITM }}$ and $\mathcal{L}_{\mathrm {MLM }}$, we now optimize the objective function for the integration of EPIC and METER as
\begin{align}
    \mathcal{L}_{\text {total}}&=\mathcal{L}_{\mathrm {ITM }}+\mathcal{L}_{\mathrm {MLM }} +\lambda \mathcal{L}_{\mathrm {ITC }}+\mathcal{L}_{\mathrm {GEN}},
\end{align}
where $\lambda$ is a hyperparameter to balance the ITC loss and losses in METER. 

\section{Improving CMLM using Saliency-based Masking}

The proposed saliency-based masking can also be applied to CMLM to alleviate the modality bias problem. 
After we obtained the masking positions $\mathcal{M}$ as described in Sec.~\ref{saliency}, we mask input sentences accordingly.
Then we conduct CMLM on masked sentences.
We term the improved CMLM as \textdagger{CMLM}.
As shown in Table \ref{tbl:improve_cmlm}, when salient tokens are masked in CMLM, the pre-trained model demonstrates improved results on downstream tasks. 
This validates our claim in Sec.~\ref{sec2} that the modality biasproblem of  prevents CMLM from learning sufficient vision-language associations. 
We also tried to replace the original CMLM in EPIC with \textdagger{CMLM} but we did not observe additional performance gain on downstream tasks.
We surmise that the improvement of \textdagger{CMLM} overlaps with that of EPIC. 

\begin{table}[!t]
    \small
	\centering	
 \setlength\tabcolsep{3pt}
	\resizebox{1.0\columnwidth}{!}{%
	\begin{tabular} {l | c  c  c  c  c  c  c  }
		\toprule	 	
	 \multirow{2}{*}{Method} & NLVR2 & \multicolumn{2}{c}{\small{Flickr30K-ft}} & \multicolumn{2}{c}{\small{Flickr30K-zs}} & \multicolumn{2}{c}{\small{MSCOCO-ft}}\\
	 & dev & TR1 & IR1 & TR1 & IR1 & TR1 & IR1 \\
	 \midrule
	\small{vanilla METER} & 79.6 & 89.2 & 76.6 & 83.2 & 67.7 & 71.0 & 52.5 \\
	\small{\textdagger{CMLM}} & \textbf{79.7} & \textbf{90.8} & \textbf{77.9} & \textbf{84.6} & \textbf{69.3} & \textbf{72.6} & \textbf{53.7} \\
	\bottomrule
	\end{tabular}
 	}
 	 \vspace{-2ex}
	\caption
	{
	\small	
		Replacing the CMLM in vanilla METER with \textdagger{CMLM} brings improvement to downstream tasks.
	}
	\label{tbl:improve_cmlm}
\end{table}

\end{document}